\documentclass[sigconf]{acmart}
\AtBeginDocument{%
 }

\copyrightyear{2023}
\acmYear{2023}
\setcopyright{rightsretained}
\acmConference[SIGIR '23]{Proceedings of the 46th International ACM SIGIR Conference on Research and Development in Information Retrieval}{July 23--27, 2023}{Taipei, Taiwan}
\acmBooktitle{Proceedings of the 46th International ACM SIGIR Conference on Research and Development in Information Retrieval (SIGIR '23), July 23--27, 2023, Taipei, Taiwan}\acmDOI{10.1145/3539618.3591879}
\acmISBN{978-1-4503-9408-6/23/07}

\acmPrice{}

\makeatletter
\gdef\@copyrightpermission{
  \begin{minipage}{0.3\columnwidth}
   \href{https://creativecommons.org/licenses/by/4.0/}{\includegraphics[width=0.90\textwidth]{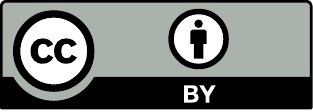}}
  \end{minipage}\hfill
  \begin{minipage}{0.7\columnwidth}
   \href{https://creativecommons.org/licenses/by/4.0/}{This work is licensed under a Creative Commons Attribution International 4.0 License.}
  \end{minipage}
  \vspace{5pt}
}
\makeatother

\usepackage{amsmath}
\usepackage{stmaryrd}
\usepackage{xcolor}
\usepackage{soul}

\usepackage{multirow}               

\usepackage{amssymb}                

\usepackage{mathtools}              
\usepackage{enumitem}               

\usepackage{subfigure}
\usepackage{makecell}

\definecolor{RoseQuartzBg}{HTML}{F7CAC9}
\definecolor{RoseQuartz}{HTML}{F5A798}
\definecolor{Serenity}{HTML}{92A8D1}
\definecolor{OrangeRed}{rgb}{1.0, 0.27, 0.0}
\definecolor{Red}{rgb}{1.0, 0.0, 0.0}
\definecolor{Turquoise}{HTML}{0F4C81}
\usepackage{pifont}
\newcommand{\cmark}{\ding{51}}%
\newcommand{\xmark}{\ding{55}}%

\newcommand{\dataset}{\textsc{Mocheg}}

\definecolor{RoseQuartzBg}{HTML}{F7CAC9}
\definecolor{RoseQuartz}{HTML}{F5A798}
\definecolor{Serenity}{HTML}{92A8D1}
\definecolor{OrangeRed}{rgb}{1.0, 0.27, 0.0}
\definecolor{Red}{rgb}{1.0, 0.0, 0.0}
\definecolor{Turquoise}{HTML}{0F4C81}
\definecolor{carnationpink}{rgb}{1.0, 0.65, 0.79}
\definecolor{darkpink}{rgb}{0.91, 0.33, 0.5}

 \setlist[itemize]{leftmargin=*}
\setlist[enumerate]{leftmargin=*}
 
\newpage
\usepackage{appendix}  
\usepackage{soul}
\usepackage{balance}

\begin{document}

\title{End-to-End Multimodal Fact-Checking and Explanation Generation: A Challenging Dataset and Models}

\author{Barry Menglong Yao}
\affiliation{%
  \institution{Virginia Tech}
  \country{Blacksburg, USA}
}\email{barryyao@vt.edu}

\author{Aditya Shah}
\affiliation{%
  \institution{Virginia Tech}
  \country{Blacksburg, USA}}
\email{aditya31@vt.edu}

\author{Lichao Sun}
\affiliation{%
  \institution{Lehigh University}
  \country{Bethlehem, USA}
}\email{lis221@lehigh.edu}

\author{Jin-Hee Cho}
\affiliation{%
 \institution{Virginia Tech}
 \country{Blacksburg, USA}
}\email{jicho@vt.edu}

\author{Lifu Huang}
\affiliation{%
  \institution{Virginia Tech}
  \country{Blacksburg, USA}
 }\email{lifuh@vt.edu}
 
\renewcommand{\shortauthors}{Barry Menglong Yao, Aditya Shah, Lichao Sun, Jin-Hee Cho, \& Lifu Huang}

\begin{abstract} 


We propose end-to-end multimodal fact-checking and explanation generation, where the input is a claim and a large collection of web sources, including articles, images, videos, and tweets, and the goal is to assess the truthfulness of the claim by retrieving relevant evidence and predicting a truthfulness label (e.g., \textit{support}, \textit{refute} or \textit{not enough information}), and to generate a statement to summarize and explain the reasoning and ruling process. To support this research, we construct \dataset{}, a large-scale dataset consisting of 15,601  claims where each claim is annotated with a truthfulness label and a ruling statement, and 33,880 textual paragraphs and 12,112 images in total as evidence. To establish baseline performances on \dataset{}, we experiment with several state-of-the-art neural architectures on the three pipelined subtasks: multimodal evidence retrieval, claim verification, and explanation generation, and demonstrate that the performance of the state-of-the-art end-to-end multimodal fact-checking does not provide satisfactory outcomes.  To the best of our knowledge, we are the first to build the benchmark dataset and solutions for end-to-end multimodal fact-checking and explanation generation. The dataset, source code and model checkpoints are available at \textcolor{darkpink}{\url{https://github.com/VT-NLP/Mocheg}}.

\end{abstract} 





\maketitle


\section{Introduction}
Misinformation has been a growing public concern in society and caused difficulty in finding reliable information online~\cite{godfrey1989misinformation,edelman2001politics}. For example, as~\citet{Islam20} shows, the misinformation about COVID-19 has widely spread and led people to distrust medical treatment and even refuse to get vaccinated. The situation has become even more complicated with the emergence of large language models, like ChatGPT~\cite{chatgpt} since they could be intentionally misused to generate misinformation~\cite{goldstein2023generative} or wrongly spread misinformation due to the hallucination issue~\cite{zhuo2023exploring}. To fight against misinformation, many fact-checking websites, such as \textit{Snopes}\footnote{\url{https://www.snopes.com/}} and \textit{PolitiFact}\footnote{\url{https://www.politifact.com/}}, have been created where journalists manually collect thousands of claims from news and social media and verify them by referring to external reliable and relevant documents. However, it is time-consuming and hard to generalize to more broad claims.

\begin{figure}[t]
    \centering
    \includegraphics[width=0.5\textwidth]{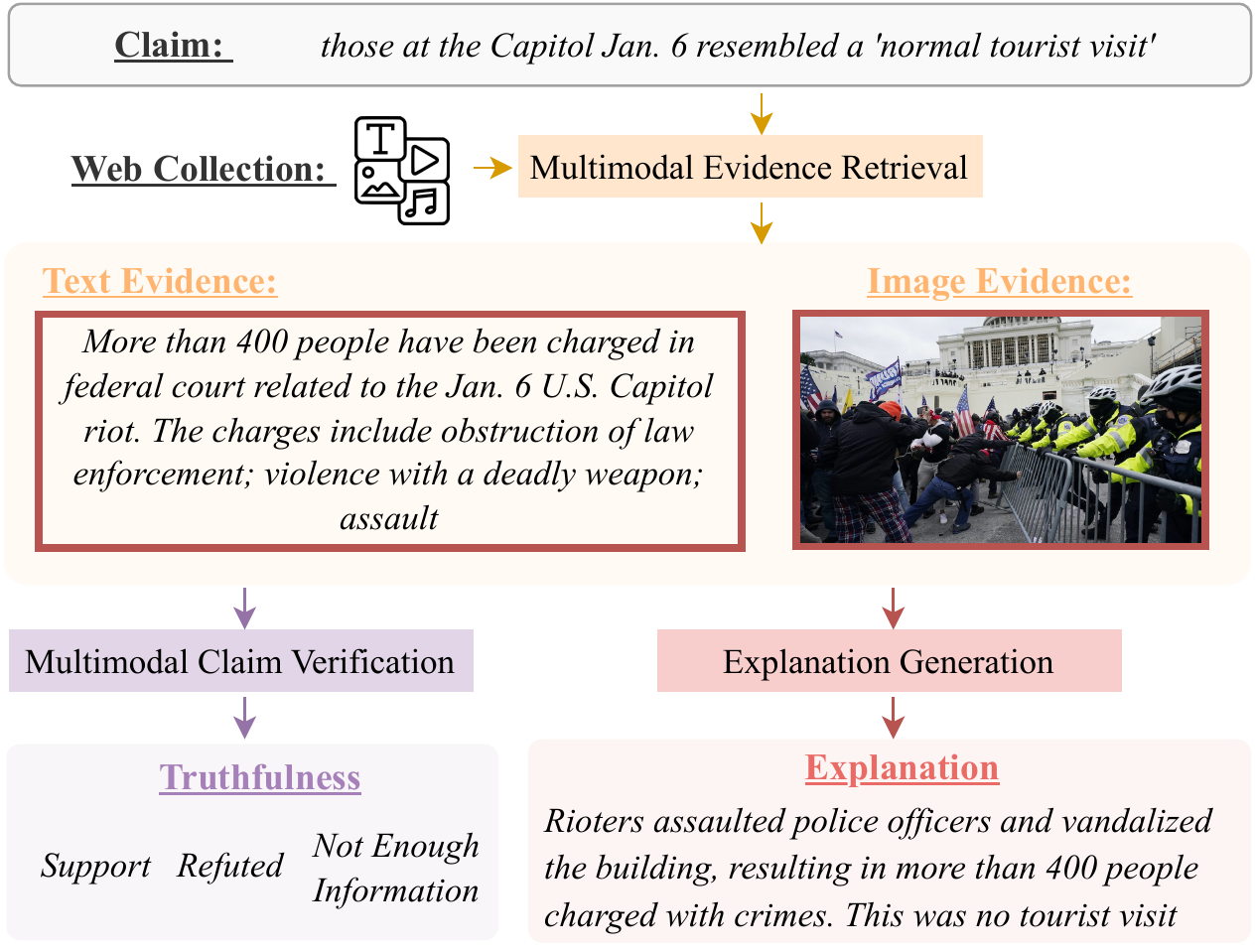}
    \vspace{-2mm}
    \caption{An example of end-to-end multimodal fact-checking and explanation generation. }
    \label{fig:example}
    \vspace{-5mm}
\end{figure}

Recently, researchers have started to investigate automatic misinformation detection and fact-checking by developing various benchmark datasets~\cite{Thorne,Wang,Shu,Nakamura2020,article} and start-of-the-art neural network architectures~\cite{Tan,Song2021,Li,Zhou}. However, we found the following limitations with the current fact-checking studies: (1) Most of them only consider text while ignoring the multi-media nature (e.g., images) of online articles, which are essential and useful to predict the truthfulness of claims. 
There are a few multimodal fact-checkinig datasets existing~\cite{Nielsen,Abdelnabi2015,Mishra2021}, however, their truthfulness labels~\cite{Mishra2021} or evidence~\cite{Nielsen,Abdelnabi2015} are automatically generated and thus cannot be guaranteed to be consistent with human judgements. 
(2) While current studies simply predict a truthfulness label, it is also necessary to provide a textual statement to explain the prediction. These explanations are vital to justify how the conclusion is reached step by step based on external evidence, and provide the public with rationale to analyze the reasoning process and share it with others. (3) Some prior studies~\cite{Wang,Zlatkova,Reis2020} assume that a short piece of evidence text is already identified, based on which the models can directly predict the truthfulness of the target claim. However, this is not realistic in practice as the claim does not come with evidence, which should be retrieved from a knowledge base or the Internet.

\begin{table*}[!htp]
 
\small
\centering
\resizebox{1.6\columnwidth}{!}{%
\begin{tabular}{l|c|c|c|c|c }
\toprule
\diaghead(-4,1){\hskip 2.3cm}%
  {\\\textbf{\small{Dataset}}}{\textbf{\small{Feature}}\\}&\textbf{Evidence Retrieval} & \textbf{Multimodal} & \textbf{Explainable Fact-checking} &  \textbf{Annotated Label} &\textbf{Annotated Evidence} \\
 \midrule
FEVER~\cite{Thorne}  &  \cmark & \xmark & \xmark&  \cmark&  \cmark
\\
Liar~\cite{Wang} & \xmark &\xmark& \xmark& \cmark & \cmark
\\
Snopes\cite{Hanselowski} & \cmark & \xmark & \xmark& \cmark & \cmark
\\
PUBHEALTH~\cite{Kotonya}  & \cmark & \xmark & \cmark& \cmark & \cmark
\\
FACTIFY~\cite{Mishra2021} & \xmark & \cmark & \xmark& \xmark & \xmark
\\
MuMiN ~\cite{Nielsen} & \xmark & \cmark   & \xmark &\cmark  & \xmark
\\
FakeNewsNet~\cite{Shu}& \cmark & \cmark & \xmark& \cmark & \xmark
\\
Fauxtography~\cite{Zlatkova} & \xmark & \cmark & \xmark&  \cmark & \xmark
\\ 
NewsBag~\cite{Jindal2020} & \xmark & \cmark & \xmark& \cmark & \xmark
\\
QProp~\cite{Barron-Cedeno2019} & \xmark & \cmark & \xmark& \cmark & \xmark
\\
TABFACT~\cite{Chen_TABFACT} & \xmark & \cmark & \xmark& \cmark & \cmark
\\
CLAIMDECOMP~\cite{Chen} & \cmark & \xmark &  \cmark& \cmark  & \cmark 
\\
MultiFC~\cite{Augenstein} & \cmark&\xmark   & \xmark& \cmark & \xmark
\\
FEVEROUS~\cite{Aly} & \cmark&  \xmark  & \xmark&  \cmark & \cmark 
\\

\midrule
\textbf{\dataset{} (Ours)} & \cmark & \cmark & \cmark&  \cmark  &  \cmark 
\\ 

\bottomrule
\end{tabular}%
}
 
\caption{Comparison between \dataset{} and other related datasets. The columns indicate whether the dataset requires automatic evidence retrieval, multimodal reasoning, or explanation generation and whether its label and evidence are annotated by a human.
}
\vspace{-8mm}
\label{tab:compare}
 
\end{table*}

To tackle these challenges, we propose end-to-end multimodal fact-checking and explanation generation, where the input consists of a claim and a large collection of web sources, including articles, images, and tweets, and the goal is to automatically retrieve information sources relevant to the claim (\textit{Evidence Retrieval}), predict the truthfulness of the claim based on the relevant evidence (\textit{Claim Verification}), and generate a textual explanation to explain the reasoning and ruling process (\textit{Explanation Generation}). An example\footnote{The example is from \url{https://www.politifact.com/factchecks/2021/may/13/andrew-clyde/ridiculous-claim-those-capitol-jan-6-resembled-nor/}} is shown in Figure~\ref{fig:example}. To support this research, we introduce \dataset{}, a new benchmark dataset with 15,601 claims annotated with truthfulness labels, multimodal evidence, and ruling statements, along with a large collection of web articles and images as the evidence sources. To set up the baseline performance, we explore the state-of-the-art pre-trained vision-language models for multimodal evidence retrieval, claim verification, and explanation generation. Experimental results show that there is still a huge room for further improvements in this end-to-end multimodal fact-checking and explanation generation task. Overall, the contributions of our work are as follows:
 \begin{itemize}
\item To the best of our knowledge, this is the first study that investigates end-to-end multimodal fact-checking and explanation generation task.
\item We create the first benchmark dataset for end-to-end multimodal fact-checking and explanation generation. The baseline performance of the state-of-the-art language models demonstrates that the task is still challenging, and there is a huge space to improve.
\end{itemize}

\section{Related work}  

\paragraph{Multimodal Fake News Detection and Fact-checking:} 
Most previous benchmark datasets~\cite{Wang,Alhindi2018,Aly,Thorne,Hanselowski,Kotonya,Augenstein,KishoreShahi} for fake news detection and fact-checking are mainly based on text. As information is naturally in multi-modality, recent studies have started to take images~\cite{boididou2015verifying,Zlatkova,Shu,Nakamura2020,Jindal2020,Reis2020,Fung,RAJ202236} and videos~\cite{papadopoulou2018corpus,Rayar2022,micallef2022cross} into consideration. Many methods for multimodal fake news detection are based on cross-modality consistency checking~\cite{Tan,Zhou,Song2021,Wang2021,Abdelnabi2015,roy2021mulcob} or computing a fused representation of multimodal (textual + visual) information for final classification~\cite{Khattar2019MVAEMV,SONG2021102437,app12031093,kamboj2020multimodal}. \cite{Reis2020,Zlatkova,Nakamura2020} directly predict the truthfulness of multimodal claims without considering explicit evidence. \cite{Mishra2021,Nielsen,Abdelnabi2015} are the most related work to ours in that it considers explicit multimodal evidence. However, their labels or evidence are automatically generated without validating by humans 
while our label and evidence are annotated by fact-checking journalists. And we further provide journalists explanations regarding the truthfulness prediction. Compared with all these studies, our \dataset{} is designed for the end-to-end multimodal fact-checking and explanation generation that requires systems to automatically retrieve multimodal evidence to predict the truthfulness of each claim and generate a ruling statement to explain the reasoning and ruling process. Table~\ref{tab:compare} compares \dataset{} with mentioned datasets. 

\paragraph{Explainable Fact-Checking:} Providing explanations to the model predictions is beneficial for humans to understand the truthfulness of the claims~\cite{Guo2022,doi:10.1080/08913811.2013.843872,4158afa7f7ca457b8c1ceb6dfb0214ab,gurrapu2022exclaim,gurrapu2023rationalization}. Current explainable fact-checking studies can be divided into four categories. The first is to directly take the evidence used for claim verification as the explanation~\cite{Thorne,Alhindi2018,Hanselowski,Fan2020}. However, the evidence usually consists of several individual sentences extracted from a large collection of documents, which are not logically connected and thus might be hard for humans to interpret. The second is to incorporate external knowledge graphs to compute a set of semantic traces starting from the claim~\cite{Gad-Elrab2019}. The semantic traces can serve as explanations to justify the truthfulness of the claims. The third is to generate questions based on claims and link the claims and evidence by using these questions as a proxy~\cite{yang2022explainable,Chen,Dai2022}. Although these generated questions can improve the explainability, they may be similar or less relevant because, normally, the claim is short. The fourth is to apply natural language generation to generate a paragraph describing the reasoning process~\cite{Atanasova2020,Kotonya,Kazemi2021,Zhang,Stammbach}, which is the most interpretable to humans. Previous studies usually summarize fact-checking articles written by journalists in shorter paragraphs as explanations. In stark contrast, our work generates explanations based on the evidence that is automatically retrieved from the web, which is more realistic in practice. In addition, in our end-to-end multimodal setting, the system needs to sequentially or jointly perform all three sub-tasks, including {\em multimodal evidence retrieval}, {\em multimodal claim verification}, and {\em multimodal explanation generation}.

 \vspace{-3mm}
 \section{Dataset Construction} \label{sec: dataset}
 \subsection{Data Source} 
PolitiFact and Snopes are two widely used websites to fight against the spreading of misinformation, where journalists are asked to manually check and verify each claim and write a ruling article to share their judgment. Considering this, we use these two websites as the data sources\footnote{We have obtained permission from both Snopes and Politifact to publish the data for the research purpose.}. Specifically, we develop scripts based on~\cite{Hanselowski} to collect all the necessary information from these two websites, including the claims that are purely based on text, truthfulness labels, text and/or image evidence that is extracted from external articles by journalists and help determine the truthfulness of claims, evidence references that are linked to external articles/images containing the text and image evidence, and ruling articles that can explain and justify the truthfulness of the claims and can be viewed as a short summary of the various evidence. Note that, the claims were originally manually collected by the journalists of the two websites from many sources, e.g., online speeches, public statements, news articles, and social media platforms, such as Facebook, Twitter, Instagram, TikTok, and so on. The truthfulness labels, evidence, evidence references, and ruling articles are also manually provided by fact-checkers of the two websites\footnote{We illustrate the detailed fact-checking processing in Snopes and Politifact in Section. \ref{appendix_fact_checking}.}. 

Based on the evidence references, we further develop scripts to collect the articles and images that contain the evidence. Since the evidence references are linked to thousands of websites with distinct HTML templates, we utilize \textit{Boilerpipe}~\cite{kohlschutter2010boilerplate} to extract the text and \textit{newspaper}\footnote{\url{https://newspaper.readthedocs.io/en/latest/}} to obtain all image links contained in the webpages and download the images based on \textit{urllib}\footnote{\url{https://docs.python.org/3/library/urllib.html}}. Some evidence references are linked to Twitter. To collect them, we first extract the Tweet IDs from the URLs of evidence references and then apply Twitter API\footnote{\url{https://developer.twitter.com/en/docs/api-reference-index}} to collect the text and images from the corresponding Tweets. 

\subsection{Data Preprocessing}

Since fact-checking websites adjust their labels over time, the initial data contains more than 75 truthfulness labels, and some labels overlap with each other, such as ``\textit{True}'',   ``\textit{TRUE}'', and ``\textit{Status: True.}''. Also, some labels have only a few instances. For example, the label ``\textit{Labeled Satire}'' has only 23 instances in total. Considering these, we follow~\cite{Hanselowski} and map 68 of these labels into three general categories, including \textit{Supported}, \textit{Refuted}, and \textit{NEI} (\textit{Not Enough Information}). We remove the claims that are annotated with other labels. In this way, each claim is just assigned one of the three target labels.
The initial dataset also contains a lot of advertisement images. To clean the dataset, we design several rules, including (1) removing an image if its name contains any of the keywords, including ``-ad-'', ``logo'', ``.gif'', ``.ico'', ``lazyload'', ``.cgi'', ``Logo'', `` .php'', ``icon'', ``Bubble'', ``svg'', ``rating-false'', ``rating-true'', ``banner'', ``-line'', or its size is smaller than 400 $\times$ 400; (2) removing a claim if we can not crawl any evidence or the ruling article; (3) for each ruling article, there is usually a paragraph starting with ``\textit{Our ruling}'' or ``\textit{In sum}'' which summarizes the whole ruling and reasoning process to achieve the fact-checking conclusion, thus we use this paragraph as the target explanation. As a result, we collect 15,601 claims with 33,880 text evidence, where each piece of text evidence is an individual paragraph extracted from a particular evidence reference article and 12,112 image evidence\footnote{Among the 15,601 claims, 19\% of them have tweets as evidence while the remaining 81\% only use other sources such as news articles or government reports as evidence. Note that the image and text evidence may be from separate sources with no clear association.}. 
Based on the evidence references, we finally collect 91,822 articles and 122,246 images which are further combined to form a constant collection of web resources for the evidence retrieval task. Within the web source collection, only 30\% (27,566 out of 91,822) of articles and 10\% (12,112 out of 122,246) of images contain the evidence of claims, making the evidence retrieve task realistic and challenging enough. 

\subsection{Task Definition}

We name the dataset \dataset{} and propose End-to-End\footnote{The end-to-end setting in our fact-checking task means it starts with only the claim and goes through the evidence retrieval, claim verification, and explanation generation, which is almost the complete pipeline for a journalist to do fact-checking in real life.} Multimodal Fact-Checking and Explanation Generation, with three subtasks\footnote{Note that we don't consider claim extraction as a subtask as all the input claims are considered worthy of being checked.}:



\noindent\textbf{Task 1. Multimodal Evidence Retrieval:} Given a claim and a collection of web sources containing both documents and images, the {\em Evidence Retrieval} task is to determine which paragraphs contained in the documents and images are related to the claim and can be further used to determine the truthfulness of the claim.

\begin{figure*}[t]
    \centering
        \includegraphics[width=0.95\linewidth]{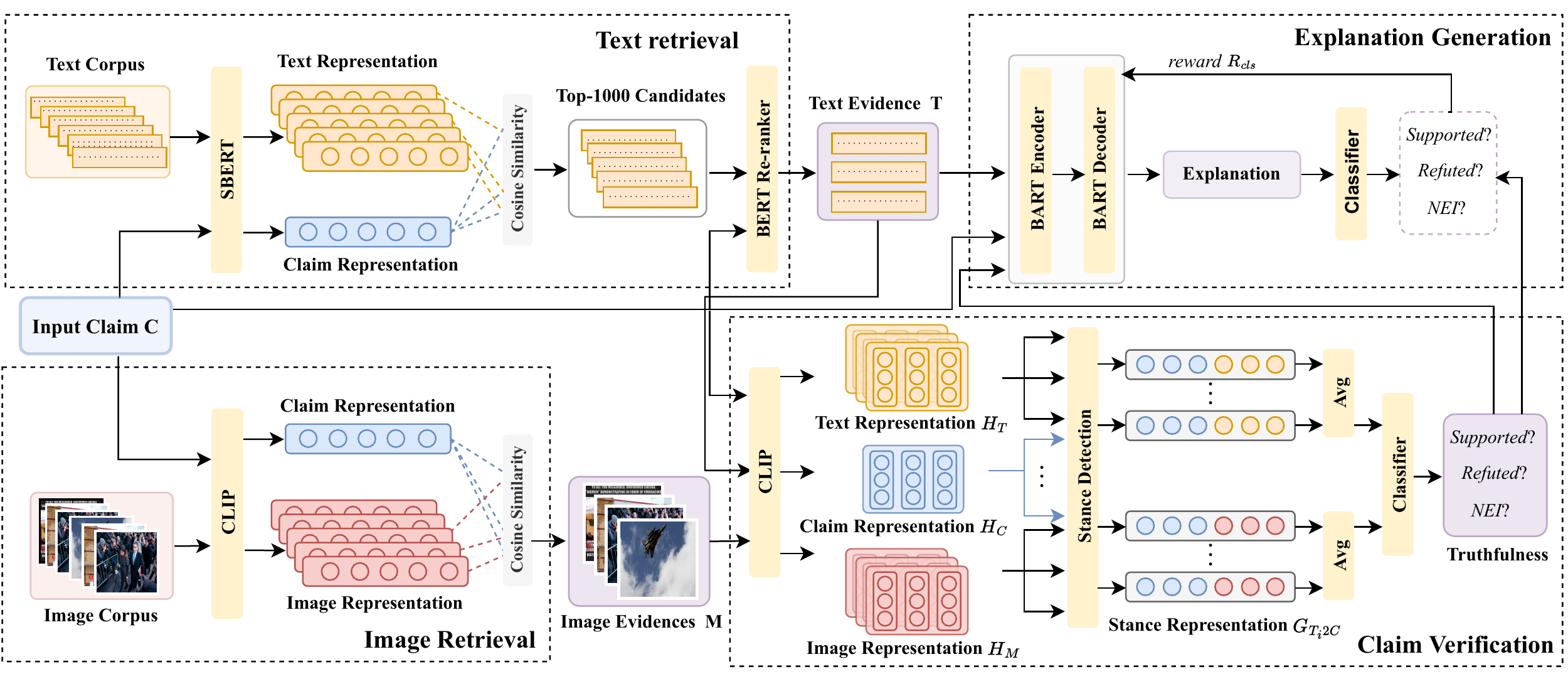}
        \vspace{-2mm}
    \caption{Overview of framework, consisting of a text evidence retrieval module (top left), an image evidence retrieval module (bottom left), a claim verification module (bottom right), and an explanation generation module (top right).}
    \label{fig:complete module}
    \vspace{-4mm}
\end{figure*}

\noindent\textbf{Task 2. Multimodal Claim Verification:} Based on the text and image evidence retrieved in Task 1, the {\em Multimodal Claim Verification} task is to predict the truthfulness (\textit{Supported}, \textit{Refuted}, or \textit{NEI}) of the claim. As both the input claim and retrieved evidence may contain both text and images, this task requires cross-modal reasoning.


\noindent\textbf{Task 3. Explanation Generation:} Given an input claim, the evidence retrieved from Task 1, as well as the truthfulness predicted from Task 2, the {\em Explanation Generation} task aims to generate a paragraph that summarizes the evidence based on the predicted truthfulness label and explains the ruling process.
    
\subsection{Train / Dev / Test Split}  
We split the whole dataset into training ({\tt Train}), development ({\tt Dev}), and test ({\tt Test}) sets with the percentage of 75\%, 10\%, and 15\%, respectively. 
Table~\ref{tab:statistic} shows the detailed statistics for each split. 

\begin{table}[ht]
\small
\centering
\resizebox{0.85\columnwidth}{!}{%
\begin{tabular}{l|c|c|c }
\toprule
\textbf{Data} & \textbf{\tt Train} & \textbf{\tt Dev} & \textbf{\tt Test} \\
 \midrule
\# Claims  &  11,669  & 1,490 & 2,442
\\ 
Ave. \# Tokens in Claim & 20 & 20 & 21  \\ 
Max. \# Tokens in Claim & 81 & 77 & 89  \\ 
\midrule
\# Text evidence (Paragraphs) & 23.545&4,067&6,268
\\  
\# Image evidence & 8,927& 1,178& 2,007
\\
 \midrule 
\# \textit{Refuted} Labels & 4,542 &488& 825
\\
\# \textit{Supported} Labels &3,826& 501 & 817
\\
\# \textit{NEI} Labels &3,301 & 501 & 800
\\ 
\midrule
Ave. \# Tokens in Explanation & 132 & 90  & 105
\\
Max. \# Tokens in Explanation & 600 & 521 & 600
\\
\midrule
\# Document/Image in Collection & \multicolumn{3}{c}{91,822 / 122,246} 
\\
\bottomrule
\end{tabular}%
}
\caption{Dataset statistics of \dataset{}.}
\label{tab:statistic}
\vspace{-4mm}
\end{table}

 \vspace{-3mm}
\section{Approach}
 

To establish the baseline performance on \dataset{}, 
we design a framework for {\em End-to-End Multimodal Fact-checking and Explanation Generation}. As illustrated in Figure~\ref{fig:complete module}, it consists of three components for the corresponding sub-tasks.   


 \subsection{Evidence Retrieval} \label{subsec:evidence-retrieval}
 
To solve this task, we apply two baseline models to retrieve text and image evidence separately.

\paragraph{Text Evidence Retrieval:} The top left in Figure~\ref{fig:complete module} illustrates the approach for text evidence retrieval. Given an input claim and a document corpus, we first split each document into sentences and then apply SBERT (Sentence-BERT)~\cite{reimers-2019-sentence-bert,reimers-2020-Curse_Dense_Retrieval} to take in the input claim and a sentence from the document corpus and output their contextual representations, based on which we can further compute a cosine similarity score for each pair. Based on these similarity scores, we rank all the sentences and select the top-$1000$ as the candidate evidence. We fine-tune the SBERT based on the following InfoNCE loss~\cite{van2018representation}:
\begin{equation*}
\resizebox{0.85\columnwidth}{!}{
$\mathcal{L}(C_i, T^{p}, \mathcal{T}) =
-\log(\frac{\exp(\text{cosine}(\boldsymbol{C}_i, \boldsymbol{T}^{p}))}{\sum_{T_j\in\mathcal{T}} \exp(\text{cosine}(\boldsymbol{C}_i,\boldsymbol{T}_j))})$
}
\end{equation*}
where $T^p$ is a piece of positive evidence to a claim $C_i$, $\mathcal{T}$ contains $T^p$ and a set of other negative evidence to $C_i$. For each claim, we use the evidence of other claims in the same batch as the negative ones\footnote{In \dataset{}, there are 37 sentences that are labeled as positive evidence of two different claims, thus the probability of a text being positive evidence of two claims in the same batch is very low.}.  
$\boldsymbol{C}_i$, $\boldsymbol{T}^p$ and $\boldsymbol{T_j}$ are the sentence level representations encoded from SBERT. In this work, we use bold symbols to denote vector representations.

We further apply a re-ranking model based on BERT~\cite{devlin2018bert}, which encodes each pair of the input claim and a candidate evidence sentence and outputs a score based on a linear classification layer. Based on these scores, we further rank all the candidate evidence and select the top-$K$ as the text evidence. The BERT-based re-ranking model is pre-trained on the MS MARCO Passage Ranking dataset~\cite{bajaj2016ms} which is designed for text retrieval.
\paragraph{Image Evidence Retrieval:} As shown in the bottom left of Figure~\ref{fig:complete module}, given an input claim and the image corpus, we use CLIP~\cite{radford2021learning} as the encoder to learn an overall representation for the claim and a representation for each image, then compute the cosine similarity between each image and the input claim. We sort all the images in the corpus based on the cosine similarity scores and take the top-$K$ as the candidate image evidence. We fine-tune CLIP based on the same InfoNCE loss as text evidence retrieval. Note that, during inference, we always retrieve top-$K$ text and image evidence respectively though it's possible that there is no image or text evidence contained in the background corpus.

  \subsection{Claim Verification}
Based on the text and image evidence, we further design a claim verification approach to predict the truthfulness of each input claim, which is shown in the bottom right of Figure~\ref{fig:complete module}. 


\paragraph{Encoding with CLIP:} We formulate an input claim as $C=\{c_0,c_1,...,c_n\}$, a piece of text evidence as $T_i=\{t_{i0},t_{i1},...,t_{is}\}$, a piece of image evidence as $M_j=\{m_{j0},m_{j1},...,m_{jq}\}$, where $c_k$ denotes the $k$-th token of the claim, $t_{ik}$ is the $k$-th token of the $i$-th text evidence $T_i$, and $m_{jk}$ is the $k$-th patch of the $j$-th image evidence $M_j$. Given a claim $C$ and its 
text evidence $\{T_{0},T_{1},...\}$ and image evidence $ \{M_{0},M_{1},...\}$, 
we concatenate them as an overall sequence $\{C, T_0, T_1, ..., M_0, M_1 ...\}$ and feed it into CLIP to obtain their contextual representations:
\begin{gather*}
\boldsymbol{H}_{C}=\{\boldsymbol{h}_{c_0},\boldsymbol{h}_{c_1},\ldots,\boldsymbol{h}_{c_n}\}, \\  
\boldsymbol{H}_{T_i}=\{\boldsymbol{h}_{t_{i0}},\boldsymbol{h}_{t_{i1}},\ldots,\boldsymbol{h}_{t_{is}}\}, \\ 
\boldsymbol{H}_{M_j}=\{\boldsymbol{h}_{m_{j0}},\boldsymbol{h}_{m_{j1}},\ldots,\boldsymbol{h}_{m_{jq}}\}.
\end{gather*}

 

\begin{figure}[h]
    \centering
    \includegraphics[width=0.5\textwidth]{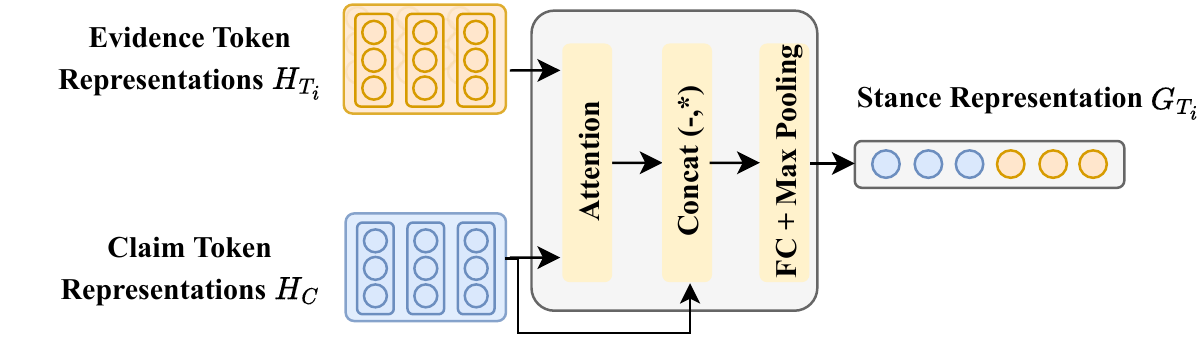}
    \vspace{-6mm}
    \caption{Stance Detection 
    }
    \vspace{-3mm}
    \label{fig:stance detection layer}
\end{figure}

\paragraph{Stance detection:} We then pair each piece of evidence with the input claim and detect the stance of the evidence towards the claim. As Figure~\ref{fig:stance detection layer} describes, taking text evidence as an example, we first compute an attention distribution between the claim and the evidence by using $\boldsymbol{H}_{C}=\{\boldsymbol{h}_{c_0},\boldsymbol{h}_{c_1},...,\boldsymbol{h}_{c_n}\}$ as query, $\boldsymbol{H}_{T_i}=\{\boldsymbol{h}_{t_{i0}},\boldsymbol{h}_{t_{i1}},\ldots,\boldsymbol{h}_{t_{is}}\}$ as key and value to compute cross attention and obtain an updated claim representation $\boldsymbol{H}_{T_{i}2C}=\{\boldsymbol{h}_{\tilde{c}_0},\boldsymbol{h}_{\tilde{c}_1},\ldots, \boldsymbol{h}_{\tilde{c}_{i}}, \ldots, \boldsymbol{h}_{\tilde{c}_n}\}$ where $\boldsymbol{h}_{\tilde{c}_{i}}$ is defined by:
\begin{gather*}
  \boldsymbol{h}_{\tilde{c}_{i}}=\text{Softmax}(\boldsymbol{h}_{c_{i}}\cdot\boldsymbol{H}^{\top}_{T_i})\cdot\boldsymbol{H}_{T_i}  
\end{gather*}

We then fuse the updated claim representation $\boldsymbol{H}_{T_{i}2C}$ with its original representation $H_C $ by two arithmetic operations, subtraction (-) and multiplication (*), which work best as comparison functions in~\cite{wang2016compare}, and obtain the stance representation $\boldsymbol{G}_{T_{i}2C}$ of evidence $T_i$ towards the claim $C$ based on max pooling.
\begin{gather*}
\boldsymbol{\tilde{G}}_{T_{i}2C}  = \sigma([\boldsymbol{H}_{T_{i}2C}\boldsymbol{H}_C:\boldsymbol{H}_{T_{i}2C}-\boldsymbol{H}_C]\cdot\boldsymbol{W}_a+\boldsymbol{b}_a),   
\\
\boldsymbol{G}_{T_{i}2C}  =  \text{Max\_Pooling}(\boldsymbol{\tilde{G}}_{T_{i}2C}),
\end{gather*}
where $[:]$ denotes concatenation operation, $\boldsymbol{W}_a$ and $\boldsymbol{b}_a$ are learnable parameters for aggregating the representations, and $\sigma$ denotes a LeckyReLU activation function. 


\paragraph{Prediction:} As we have multiple text and image evidence, we further compute the average of the stance representations of all text evidence and image evidence, respectively, to obtain $\boldsymbol{G}_{T2C} =  \text{Mean\_Pooling}(\boldsymbol{G}_{{T_i}2C})$ and $\boldsymbol{G}_{M2C} =  \text{Mean\_Pooling}(\boldsymbol{G}_{{M_j}2C})$. 
We then concatenate the overall stance representations\footnote{Since the evidence in our corpus is annotated by journalists on Politifact and Snopes, we assume the evidence is reliable and fuse the stance of evidence to the claim to predict the truthfulness. We leave it as a future work to check the trustworthiness of evidence.
} $\boldsymbol{G}_{T2C}$ and $\boldsymbol{G}_{M2C}$ obtained from both modalities to predict the truthfulness label and optimize the claim verification approach based on the cross-entropy objective:
\begin{gather*}
\boldsymbol{\hat{y}}_{cls} = \boldsymbol{W}_h^{\top}\cdot[\boldsymbol{G}_{{T}2C}: \boldsymbol{G}_{{M}2C}]+\boldsymbol{b}_h , 
\\
\mathcal{L}(y_i|C) =-\log(\frac{\exp(\boldsymbol{\hat{y}}_{cls,i})}{\sum_{j=0}^2 \exp(\boldsymbol{\hat{y}}_{cls,j})})
\end{gather*}
where $\boldsymbol{\hat{y}_{cls}}$ denotes the probabilities over all possible labels. $y_i$ is the truthfulness label of claim $C$. During training, we fix the parameters of CLIP while tuning all the other parameters.

\subsection{Explanation Generation} \label{subsec:explanation-generation}


To justify the truthfulness prediction, we further generate a ruling statement by considering the input claim, the predicted truthfulness label as well as the text evidence. The top right of Figure~\ref{fig:complete module} illustrates the overall architecture for explanation generation. 


    

Specifically, given an input claim $C$, its truthfulness label $y_C$, and text evidence $\{T_1, T_2, \ldots\}$, we concatenate them into an overall sequence $X$ with a separator \texttt{</s>}. Then we feed this sequence as input to BART~\cite{lewis2019bart}, which is a state-of-the-art pre-trained sequence-to-sequence model, to generate a ruling statement $S=\{s_1, s_2, \ldots, s_q\}$. During training, we use the gold truthfulness label of each claim as input.  During the evaluation, we use the truthfulness label predicted by the claim verification model. The training objective is to minimize the following negative log-likelihood based on the gold ruling statement $\tilde{S}=\{\tilde{s}_1, \tilde{s}_2, \ldots, \tilde{s}_q\}$:
\begin{equation*}
\mathcal{L}_{g}=-\sum_i\log(p(\tilde{s}_i|\tilde{s}_{1:i-1},X;\phi))
\end{equation*}

To ensure the generated ruling statement is consistent with the truthfulness label of the claim, we apply a truthfulness reward and optimize the generation model with reinforcement learning (RL)~\cite{Lai2021}. Specifically, we pre-train a truthfulness classification model based on BERT~\cite{Devlin2019}, which takes the ruling statement as input and outputs a confidence score for each truthfulness label. We use the difference between the confidence score of the correct label and the score of the wrong labels as the reward $R_{cls}$:
\begin{gather*}
R_{cls} =\boldsymbol{p}(y_C|S) \; \; - \sum_{y_j\neq y_C, y_j\in Y}\boldsymbol{p}(y_j|S), \\
\boldsymbol{p}(y|S) =\text{Softmax}(\text{BERT}_{\theta}(S)),
\end{gather*}
where $y_C$ is the gold truthfulness label of $C$, $Y$ is the target label set, and $S$ is the generated ruling statement. 

We then apply the reward $R_{cls}$ for policy learning, and the policy gradient is computed as:
\begin{equation*}
\nabla_{\phi}{\mathcal{J}({\phi})} =\mathbb{E}[ R_{cls}\cdot\nabla_{\phi}\sum_i\log(\boldsymbol{p}(s_i|s_{1:i-1},X;\phi))],
\end{equation*}
where $X$ is the concatenated sequence of the input claim, its truthfulness label, and text evidence, and $\phi$ denotes the model parameters.
\section{Experiments}  

\subsection{Evidence Retrieval} \label{subsec:ex-ER}
For each claim, we retrieve the top-$K$ text and image evidence from the corresponding text and image corpus and
evaluate the retrieval performance based on Precision, Recall, NDCG~\cite{jarvelin2002cumulated}, MAP (Mean Average Precision), and S-Recall (Similarity-based Recall) scores. In S-Recall, it first computes a recall score for each gold text or image evidence based on the highest cosine similarity score between it and all retrieved text or image evidence, while each piece of evidence is represented with a vector learned from SBERT or CLIP. We use the average recall of all gold evidence as the S-Recall.


\begin{table}[!htp]
\centering
\footnotesize
\resizebox{\columnwidth}{!}{
\begin{tabular}{l|c|c|c|c|c|c}
\toprule
  \textbf{Media}  & \textbf{K}&\textbf{Rec@K }&\textbf{Pre@K} &\textbf{NDCG }&\textbf{MAP }&\textbf{S-Rec }  \\
\midrule
 Image & 5& 17.01& 4.71 & 13.81 & 11.93 &  68.22 \\
  Image & 10&  21.44&  3.02& 15.32 &  12.58&  71.85 \\
 \midrule
 Text w/o Re-ranking &5  &15.67 & 12.20 & 19.23 &13.61  & 52.42  \\
 Text w/o Re-ranking &10& 19.40& 8.16 & 19.60 &13.02  & 55.77 \\
 \midrule
 Text &5&19.72  & 14.92&23.66  & 14.34 & 54.57 \\
 Text & 10& 23.99& 9.79 & 24.09  &15.34  & 58.28  \\
\bottomrule
\end{tabular}%
}
\caption{Performance of text and image evidence retrieval. (\%). \textit{Pre} denotes \textit{Precision} while \textit{Rec} means \textit{Recall}.}
 \label{tab:widgets1}
\vspace{-7mm}
\end{table}

We show the performance of text and image evidence retrieval on the test set of \dataset{} in Table~\ref{tab:widgets1}. We can see that the performance of both image and text evidence retrieval is low, indicating the difficulty of both tasks. Taking text evidence retrieval as an example, the model needs to retrieve 2 pieces of text evidence on average for each claim from a collection of 2,792,639 sentences, which is very challenging. 
Also, the proposed evidence retrieval is based on semantic matching. However, in many cases, it is more important to find evidence that is relevant to the claim but describes different aspects or is against the claim, especially for refuted claims. For example, given an input claim, \textit{``H.R. 6666 provides \$100 billion to entities that perform COVID-19 testing but prohibits them from allowing any non-vaccinated persons into their facilities.''} the retrieval model missed an important piece of evidence \textit{``No provision in this bill would make testing or quarantining mandatory.''}. This is against the claim and has lower similarity compared with the retrieved text \textit{``It would provide \$100 billion to organizations that do COVID-19 testing or contact tracing or that provide services to people who are isolated at home.''}. In addition, for many claims, their evidence come from the comprehension of long paragraphs instead of several sentences. Although our approach successfully retrieves several relevant sentences, they are insufficient to cover all the background and indicate the truthfulness of the claims.

\subsection{Claim Verification}

For claim verification, we first design two common baselines: (1) \textit{Majority Label}, which predicts the majority label (i.e., \textit{Refuted}) in the \texttt{Training} set for all the claims in the \texttt{Test} set; and (2) \textit{Average Similarity}, which computes average cosine similarity between the target claim and all the gold text and image evidence based on their embeddings learned from CLIP. If the average similarity is higher than $\alpha_1\in\{0.5, 0.6, 0.7, 0.75, 0.8\}$, predict it as \textit{Supported}; if the average similarity is lower than $\alpha_2\in\{0.2, 0.3, 0.4, 0.5, 0.6, 0.65, 0.7\}$ and $\alpha_2<\alpha_1$, predict it as \textit{Refuted}; otherwise, predict it as \textit{NEI}. We search for the best value of $\alpha_1$ and $\alpha_2$ on the \texttt{Development} set and then apply them to the \texttt{Test} set. 
We then adapt Pre-CoFactv2~\cite{du2023team}, a multimodal fact-checking model which achieves state-of-the-art results at the
Factify 2 challenge~\cite{factify2023} at AAAI 2023\footnote{\url{https://aiisc.ai/defactify2/}}, to be the third baseline model. As there is very little existing work on multimodal fact-checking,
we further adapt SpotFakePlus~\cite{singhal2020spotfake+}, a multimodal fake news detection approach, to our fact-checking task, by using their model to compare the consistency of input claim and image evidence and adding a new component to check the consistency of input claim and text evidence\footnote{Most of existing multimodal fake news detection studies aim to detect fake news by comparing the consistency between news text and news image or between the news articles and external knnowledge graphs, thus cannot be directly applied to fact-checking task.}.   
As shown in Table~\ref{tab:widgets2}, \textit{Majority Label} and \textit{Average Similarity} yield a performance score that is close to a random baseline, while Pre-CoFactv2 and SpotFakePlus underperform our approach, demonstrating that \dataset{} does not contain any label distribution bias and cannot be easily solved simply by comparing the semantics between claims and evidence. 


\begin{table}[h]
\centering
\resizebox{0.36\textwidth}{!}{%
\begin{tabular}{l|c}
\toprule
\textbf{Setting} & \textbf{F-score (\%)} \\ 
\midrule
Majority Label & 33.78\\
Average Similarity (Gold Evidence) & 32.72\\ 
Pre-CoFactv2~\cite{du2023team} (Gold Evidence) &  47.17\\
SpotFakePlus~\cite{singhal2020spotfake+} (Gold Evidence) & 44.11 \\

\midrule
w/o Evidence & 39.93  \\
\midrule
w/ Text Evidence (Gold) & 47.54 \\ 
w/ Image Evidence (Gold) &45.62  \\ 
w/ Text and Image evidence (Gold) & 50.78 \\ 
\midrule
w/ Text Evidence (System) &    42.79\\ 
w/ Image Evidence (System) &  40.91    \\ 
w/ Text and Image evidence (System) & 44.06\\ 
\midrule
Human w/o Evidence & 20.00 \\
Human w/ System Evidence & 62.00\\
Human w/ Gold Evidence &70.00 \\
\bottomrule
\end{tabular}%
}
\caption{\label{tab:widgets2} Performance of claim verification. \textit{Gold Evidence} denotes \textit{gold text and image evidence} while \textit{System Evidence} means \textit{system-retrieved text and image evidence}.
}
\vspace{-6mm}
\end{table}

\begin{figure*}[!htp]
    \renewcommand{\thefigure}{4}
    \centering
    \includegraphics[width=0.9\linewidth]{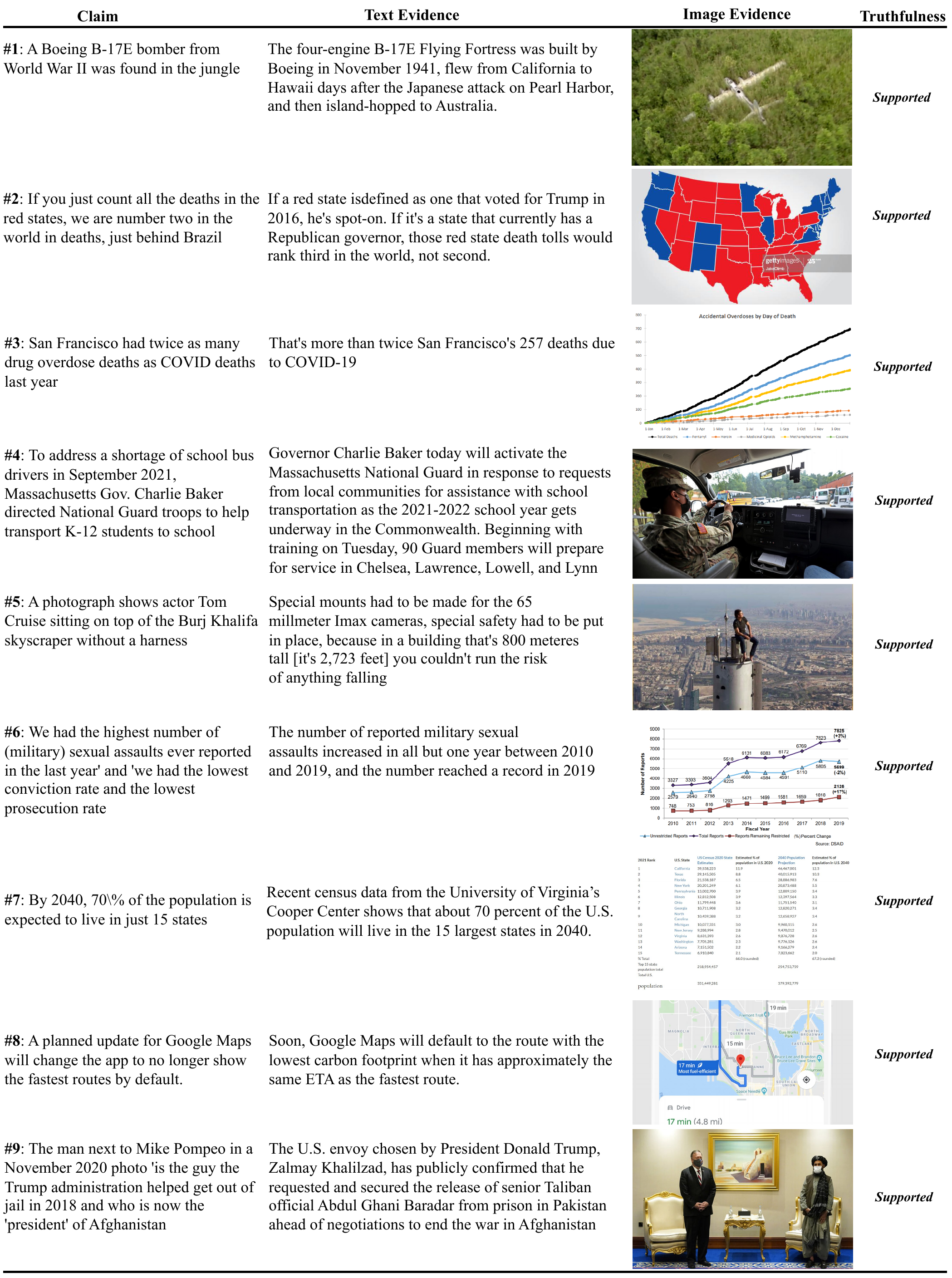}
    \caption{Examples of Multimodal Fact Checking. The \textit{Truthfulness} column shows gold labels. 
    }
    \label{fig:discussion example}
\end{figure*}

To evaluate the impact of each type of evidence to claim verification, we design ablated models of our approach by considering the text evidence only, image evidence only, or no evidence. In addition, we compare its performance based on the system-retrieved evidence and the gold evidence to show the impact of evidence retrieval. As shown in  Table~\ref{tab:widgets2}, without considering any evidence, the model can still outperform the majority based baseline on claim verification due to the fact that some claims, such as ``\textit{Paying taxes is optional!!},'' contain obvious clues or are against common sense 
so that the model can directly predict the truthfulness based on the claim itself. By adding text and/or image evidence, the performance of claim verification can be boosted, proving the usefulness of the evidence. The text evidence provides more significant gain than image evidence due to two reasons: (1) for about 32\% of the claims (787 out of 2,442) in the \texttt{Test} set, they only have text evidence without any associated image evidence. However, our approach always returns the top-5 most relevant images as evidence, introducing noises; (2) Texts usually carry more information than images. 
However, we also observe many examples that the image evidence complements the text evidence. 
For example, for claim \#1 \textit{A Boeing B-17E bomber from World War II was found in the jungle} in Figure~\ref{fig:discussion example}, its image evidence plays a crucial role in confirming \textit{the aircraft was found in the jungle}.


Finally, we also set up a human performance for claim verification by randomly sampling 50 claims and asking two annotators to label truthfulness by providing gold evidence, system evidence, or no evidence, which reach a Fleiss $\kappa$ score~\cite{fleiss1971measuring} of 0.67, 0.59, and 0.42, respectively. We take a human prediction as true only if both of the two annotators provide the true label. 
As we can see, there is still a significant gap between machine and human performance.

\subsection{Explanation Generation} \label{subsec:ex-explanation-generation}

\begin{table*}[t]
\centering
\resizebox{0.8\columnwidth*2}{!}{%
\begin{tabular}{l|c|c|c|c|c|c  }
\toprule
\textbf{Setting} & \textbf{Model} & \textbf{ROUGE 1} & \textbf{ROUGE 2} & \textbf{ROUGE L} & \textbf{BLEU} & \textbf{BERTScore}   \\
\midrule
Gold Evidence ORACLE   & -  &  40.22 &23.80  &25.97 &20.03  &
86.82 \\
Gold Evidence LEAD-3  & -  &  32.10 &16.97  &22.17 &8.41  &
86.77
\\
Gold Evidence w/o Generation  & -  &  37.71 &21.70  &25.62 &22.56  &
87.20 \\
System Evidence w/o Generation  & -  & 28.69  &  9.93&17.18 & 7.38 &
83.95\\
\midrule
Gold Evidence + Gold Truthfulness & BART-large & 45.51  &  27.37& 35.41&  21.84& 89.05\\ 
Gold Evidence + System Truthfulness & BART-large& 43.87 & 26.37&  34.10& 20.86&88.87\\
System Evidence + Gold Truthfulness & BART-large  & 35.53  & 17.46 & 26.05&10.95 &  87.01 \\ 
System Evidence + System Truthfulness & BART-large &33.88 &16.51 &  24.83&  10.08&86.95 \\

 
\bottomrule
\end{tabular}%
}
\caption{\label{tab:widgets3}  Performance of explanation generation. (\%) }
\vspace{-6mm}
\end{table*}

We fine-tune BART based on a pre-trained \texttt{bart-large}\footnote{https://huggingface.co/facebook/bart-large} checkpoint~\cite{wolf2019huggingface} to generate the ruling statement. We use ROUGE~\cite{lin-2004-rouge}, BLEU~\cite{papineni2002bleu}, and Bertscore~\cite{Zhang2019} as the evaluation metrics. The BERT-based\footnote{https://huggingface.co/bert-base-uncased} classifier is pre-trained on the gold explanations and reaches an F-score of 87.59\%. We fix the classifier during training of the generation model. To evaluate the impact of the evidence retrieval and claim verification on explanation generation, we compare the performance of our approach based on gold evidence and/or gold truthfulness labels with the system-based evidence and truthfulness labels. Note that we only train the model based on gold evidence and truthfulness but perform inference by taking different types of evidence or truthfulness as input. Similar as~\cite{Kotonya}, we further compare our method to LEAD-3, which selects the first three sentences in evidence, and the ORACLE baseline~\cite{narayan2018dont}, which greedily select\footnote{\url{https://github.com/pltrdy/extoracle_summarization}} multiple evidence sentences that maximize the ROUGE-2 score. Table~\ref{tab:widgets3} shows the results with the following observations: (1) Without generation, the explanation is directly from the concatenation of all the text evidence. The explanation may contain all the necessary information but is not interpretable to humans as the sentences are not connected coherently or logically; (2) Evidence retrieval has a more significant impact on explanation generation than claim verification.  This is reasonable because the evidence carries most of the content in the explanation and truthfulness is usually implied when comparing the evidence and the input claim. (3) The explanation in our corpus is pretty abstractive, as corroborated by the low performance of ORACLE baseline, which is the upper bound of extractive summarization, and LEAD-3 baseline. 

\subsection{Implementation Details}
We use 2 Quadro RTX 8000 to run our experiments. The retrieval models cost 15 GB and are trained for about 20 runs with a batch size of 256. The claim verification models cost 3 GB and are trained for about 50 runs with a batch size of 128. The explanation generation model costs 45 GB and is trained for about 30 runs with a batch size of 10. We use grid search to tune the hyperparameters: for evidence retrieval, the learning rage $\in\{10^{-5}, 10^{-6}, 10^{-7}\}$ and batch size $\in\{256, 480, 512\}$; for claim verification, the learning rage $\in\{10^{-1}, 10^{-2}, 10^{-3}, 10^{-4}\}$ and batch size $\in\{64, 128, 256, 512, 1024, 2048\}$; for explanation generation, the learning rage $\in\{5\times10^{-2}, 5\times10^{-3}, 5\times10^{-4}, 5\times10^{-5}\}$ and batch size $\in\{10, 12, 48, 192\}$.



\section{Remaining Challenges} 

\subsection{Claim Verification}


We randomly sample 50 claims with gold evidence that are incorrectly verified from the \texttt{Test} set and identify the following remaining challenges for multimodal fact-checking:

\vspace{1mm}
\noindent\textbf{Cross-modality Reasoning:} Both text evidence and image evidence provide complementary information to verify the truthfulness of the claims. 30\% of verification errors are due to deep cross-modality reasoning and evidence fusion. For example, for claim \#2 \textit{``'If you just count all the deaths in the red states, we are number two in the world in deaths.''} in Figure~\ref{fig:discussion example}, since there are two different definitions for the red state, the model needs to refer to the image map to confirm the mentioned states.



\vspace{1mm}
\noindent\textbf{Cross Document/Sentence Reasoning:} 30\% of verification errors are due to the reasoning across multiple pieces of textual evidence or across multiple sentences. For example, given the claim \textit{`The Biden administration's American Jobs Plan will be 'the biggest non-defense investment in research and development in the history of our country.''}, the model needs to first know the current largest investment is \$11 billion by referring to evidence \textit{``The largest increase in research and development came in 1964, and totaled \$11 billion''}, and then refer to another piece of evidence \textit{``experts say the plan is likely to far exceed \$11 billion in spending on research and development.''} to understand that the Plan will exceed \$11 billion.

\vspace{1mm}
\noindent\textbf{Deep Visual Understanding:} For 6\% of wrongly predicted claims, their image evidence is charts, tables, or even maps. The current visual understanding techniques, such as CLIP, cannot deeply understand the content and semantics of such images. For example, given claim \#3 \textit{``San Francisco had twice as many drug overdose deaths as COVID deaths last year''} in Figure~\ref{fig:discussion example}, to determine the truthfulness of this claim, the model needs to obtain the number of drug overdose deaths from the image.


\vspace{1mm}
\noindent\textbf{Other Complex Reasoning:} Many claims also require various types of complex reasoning, such as mathematical calculation (4\% of errors) and commonsense (8\% of errors). For instance, the model needs to understand that ``\textit{29,000 recipients}'' plus ``\textit{12,700 recipients}'' is ``\textit{41,700 recipients}'', ``\textit{from 2019 to 1998}'' is ``\textit{22 years}'', ``\textit{there are fifty states in US}''. In addition, the model has difficulty in dealing with claims (12\% of errors) that are partially supported or refuted. For example, for the claim \textit{``Since 2010, student debt has increased by 102\% and real wages have fallen by over 8\%.''}, it's true that \textit{``student debt has increased by 102\%''} but the \textit{``real wages have fallen by over 8\%''} is not correct. 

\subsection{Explanation Generation}
We also sample 50 system-generated explanations and analyze their error types as follows.

\vspace{1mm}
\noindent\textbf{Limited Encoding and Decoding Length:} Our approach is based on pre-trained language models, such as BERT and BART, which can only encode or decode a limited length of the sequence. In our dataset, some evidence and ruling statements exceed the maximal length. For those cases, we truncate the sequence and lose part of the information. 

\vspace{1mm}
\noindent\textbf{Missing Evidence:} As we construct the evidence source collection based on the evidence links listed on Snopes and PolitiFact, some evidence used in the ruling statements is not included. For example, given the claim ``\textit{By revoking the Keystone pipeline permit, Biden is destroying 11,000 jobs}'' the gold explanation contains the information ``\textit{A 2014 report found that the company would need only 50 employees to maintain the Keystone XL pipeline}'' which is not covered in any of the background documents. In addition, our current explanation generation approach only leverages text evidence while image evidence can also provide complementary information.

\vspace{1mm}
\noindent\textbf{Logical Coherence: } One critical challenge for explanation generation is to determine the logical connection among the evidence sentences and organize them coherently, a common issue in long-form text generation~\cite{hu2022mocha,hu2022planet}. For example, given the claim, ``\textit{A new, independent study found that at least 55 of our largest corporations used various loopholes to pay zero federal income tax in 2020.}'', our explanation generation approach fails to correctly organize the following two evidence: ``\textit{many of the relevant provisions are deliberate attempts to set incentives}'' and ``\textit{Some critics say the financial disclosures used to compile the report are imperfect estimates}.''

\section{Conclusion}  

We created~\dataset{}, an end-to-end multimodal fact-checking and explanation generation benchmark dataset which consists of 15,601 claims annotated with truthfulness labels, together with 33,880 text evidence, 12,112 image evidence as well as explainable statements. We explore the state-of-the-art neural architectures to set up the baseline performance on three sub-tasks (i.e., multimodal evidence retrieval, claim verification, and explanation generation).  Our experimental results show that the performance of all three sub-tasks is still far from enough. 
For future work, an obvious next step is to explore more advanced techniques to improve the three sub-tasks and deep visual understanding. Furthermore, open-domain fact-checking is another promising direction to detect hallucination errors in large language models like ChatGPT~\cite{chatgpt}. In the open-domain setting, evaluating the trustworthiness of evidence will play a critical role.


\section{Ethical Statement}
\label{copyright}



For dataset release, we have obtained permission from both Snopes and Politifact to publish the data for the research purpose. Our dataset is licensed under the CC BY 4.0\footnote{\url{https://creativecommons.org/licenses/by/4.0/}}, while the associated codes to \dataset{} for data crawler and baseline are licensed under Apache License 2.0\footnote{\url{https://www.apache.org/licenses/LICENSE-2.0}}. Our dataset contains 2,916 tweets. In accordance with
the Twitter developer terms\footnote{\url{https://developer.twitter.com/en/developer-terms/more-on-restricted-use-cases}}, we will only share the Twitter IDs and scripts to crawl tweets based on Twitter API. 
Our work can be used to predict the truthfulness of various claims in the web and stop the spread of misinformation. Our dataset does not use features or label information about sensitive personally identifiable information, like individual names. Since our dataset contains internet claims, some claims may be offensive. However, we crawl the articles from some reputational fact-checking websites, like Politifact and Snopes, to decrease the possibility of offensive content.

\label{appendix_fact_checking}
Given the importance of fact-checking in secular societies, we introduce the fact-checking process of Snopes and Politifact to show how our data sources reduce bias.
According to Politifact\footnote{ \url{https://www.politifact.com/article/2018/feb/12/principles-truth-o-meter-politifacts-methodology-i/}} and Snopes\footnote{\url{https://www.snopes.com/transparency/}}, they always attempt to contact the person, website, or organization that made the statement they are fact-checking. They will have consultations with a variety of expertise. They seek direct access to government reports, academic studies, and other data. They also have one to two rounds of reviews. Finally, they will accept the error correction from the public and mark the corrected articles. According to Politifact, PolitiFact journalists avoid the public expression of political opinion and public involvement in the political process to set their own opinions aside as they work to uphold principles of independence and fairness. 23 of 36 journalists are women. According to Snopes, members of their editorial staff are precluded from donating to or participating in political campaigns, political party activities, or political advocacy organizations. 6 of 10 journalists are women.


\begin{acks}
This research is based upon work supported by U.S. DARPA KMASS Program \# HR001121S0034. The views and conclusions contained herein are those of the authors and should not be interpreted as necessarily representing the official policies, either expressed or implied, of DARPA or the U.S. Government. The U.S. Government is authorized to reproduce and distribute reprints for governmental purposes notwithstanding any copyright annotation therein.
\end{acks}

\bibliographystyle{ACM-Reference-Format}
\balance
\bibliography{custom}


\end{document}